\RequirePackage{amsmath}
\documentclass[runningheads]{llncs}

\usepackage{amsmath}
\usepackage{amssymb}
\usepackage{stmaryrd}
\usepackage{color}
\usepackage{url}
\usepackage{multirow}
\usepackage{float}
\usepackage{mathtools}
\usepackage[misc]{ifsym}
\usepackage[ruled,linesnumbered]{algorithm2e}
\usepackage{multicol}

\newcommand{\sect}{Section}
\newcommand{\fig}{Fig.}
\newcommand{\tab}{Table}
\newcommand{\alg}{Alg.}
\newcommand{\algln}{line}

\newcommand{\comment}[1]{\hfill$\triangleright$ #1}   
\let\oldnl\nl
\newcommand{\nonl}{\renewcommand{\nl}{\let\nl\oldnl}} 

\newcommand{\labl}{\lambda}             
\newcommand{\iterlabels}{k}             
\newcommand{\numlabels}{K}              
\newcommand{\labelset}{\mathcal{L}}     
\newcommand{\labelspace}{\mathcal{Y}}   

\newcommand{\truelabel}{y}              
\newcommand{\truelabelvect}{\vec{y}}    

\newcommand{\predlabel}{\hat{y}}        
\newcommand{\predvect}{\vec{\hat{y}}}   

\newcommand{\conf}{\hat{p}}             
\newcommand{\confvect}{\vec{\hat{p}}}   

\newcommand{\ex}{\vec{x}}               
\newcommand{\iterex}{n}                 
\newcommand{\numex}{N}                  

\newcommand{\attr}{A}                   
\newcommand{\iterattr}{l}               
\newcommand{\numattr}{L}                
\newcommand{\attrval}{x}                
\newcommand{\attrspace}{\mathcal{X}}    

\newcommand{\dataset}{\mathcal{D}}      

\newcommand{\classifier}{f}             
\newcommand{\ensemble}{F}               
\newcommand{\iterclassifiers}{t}        
\newcommand{\numclassifiers}{T}         

\newcommand{\rul}{\classifier}          
\newcommand{\body}{b}                   
\newcommand{\head}{\confvect}           

\newcommand{\loss}{\ell}                

\newcommand{\objective}{\mathcal{R}}                    
\newcommand{\approxobjective}{\widetilde{\mathcal{R}}}  
\newcommand{\regweight}{\omega}                         
\newcommand{\regterm}{\Omega}                           

\newcommand{\gradient}{g}                               
\newcommand{\gradientvect}{\vec{g}}                     
\newcommand{\hessian}{h}                                
\newcommand{\hessianmatr}{H}                            
\newcommand{\regmatr}{R}                                

\newcommand{\iterrows}{i}   
\newcommand{\itercols}{j}   

\DeclareMathOperator{\diag}{diag}           
\DeclareMathOperator{\bigo}{\mathcal{O}}    

\newcommand{\numbins}{B}            
\newcommand{\iterbins}{b}           

\newcommand{\pos}{\oplus}
\renewcommand{\neg}{\ominus}
\newcommand{\numbinspos}{B_{\pos}}      
\newcommand{\numbinsneg}{B_{\neg}}      
\newcommand{\widthpos}{w_{\pos}}        
\newcommand{\widthneg}{w_{\neg}}        
\newcommand{\maxpos}{\max_{\pos}}       
\newcommand{\minpos}{\min_{\pos}}       
\newcommand{\maxneg}{\max_{\neg}}       
\newcommand{\minneg}{\min_{\neg}}       

\newcommand{\bin}{\mathcal{B}}          
\newcommand{\mapfun}{m}                 
\newcommand{\crit}{c}                   

\newcommand{\bingradients}{{\widetilde{\gradient}}}         
\newcommand{\binhessians}{{\widetilde{\hessian}}}           
\newcommand{\bingradientvect}{{\widetilde{\gradientvect}}}  
\newcommand{\binhessianmatr}{{\widetilde{\hessianmatr}}}    
\newcommand{\binregmatr}{\widetilde{\regmatr}}              

\begin{document}

\title{Gradient-based Label Binning\\ in Multi-label Classification}
\titlerunning{Gradient-based Label Binning in Multi-label Classification}
\toctitle{Gradient-based Label Binning in Multi-label Classification}

\author{Michael Rapp (\Letter)\inst{1} \and Eneldo Loza Menc\'ia\inst{1} \and\\ Johannes F\"urnkranz\inst{2} \and Eyke H\"ullermeier\inst{3}}
\authorrunning{Rapp~et~al.}
\tocauthor{Michael Rapp, Eneldo Loza Menc\'ia, Johannes F\"urnkranz, Eyke H\"ullermeier}

\institute{Knowledge Engineering Group, TU Darmstadt, Darmstadt, Germany\\
\email{mrapp@ke.tu-darmstadt.de}, \email{research@eneldo.net}
\and
Computational Data Analysis Group, JKU Linz, Linz, Austria\\
\email{juffi@faw.jku.at}
\and
Heinz Nixdorf Institute, Paderborn University, Paderborn, Germany\\
\email{eyke@upb.de}
}

\maketitle

\begin{abstract}

In multi-label classification, where a single example may be associated with several class labels at the same time, the ability to model dependencies between labels is considered crucial to effectively optimize non-decomposable evaluation measures, such as the Subset 0/1 loss. The gradient boosting framework provides a well-studied foundation for learning models that are specifically tailored to such a loss function and recent research attests the ability to achieve high predictive accuracy in the multi-label setting. The utilization of second-order derivatives, as used by many recent boosting approaches, helps to guide the minimization of non-decomposable losses, due to the information about pairs of labels it incorporates into the optimization process. On the downside, this comes with high computational costs, even if the number of labels is small. In this work, we address the computational bottleneck of such approach --- the need to solve a system of linear equations --- by integrating a novel approximation technique into the boosting procedure. Based on the derivatives computed during training, we dynamically group the labels into a predefined number of bins to impose an upper bound on the dimensionality of the linear system. Our experiments, using an existing rule-based algorithm, suggest that this may boost the speed of training, without any significant loss in predictive performance.

\end{abstract}

\section{Introduction}
\label{sec:intro}

Due to its diverse applications, e.g., the annotation of text documents or images, \emph{multi-label classification} (MLC) has become an established topic of research in the machine learning community (see,~e.g.,~\cite{zhang2014} or \cite{gibaja2014} for an overview). Unlike in traditional classification settings, like binary or multi-class classification, when dealing with multi-label data, a single example may correspond to several class labels at the same time. As the labels that are assigned by a predictive model may partially match the true labeling, rather than being correct or incorrect as a whole, the quality of such predictions can be assessed in various ways. Due to this ambiguity, several meaningful evaluation measures with different characteristics have been proposed in the past (see,~e.g.,~\cite{tsoumakas2010}). Usually, a single model cannot provide optimal predictions in terms of all of these measures. Moreover, empirical and theoretical results suggest that many measures benefit from the ability to model dependencies between the labels, if such patterns exist in the data \cite{dembczynski2012a}. As this is often the case in real-world scenarios, research on MLC is heavily driven by the motivation to capture correlations in the label space.

To account for the different properties of commonly used multi-label measures, the ability to tailor the training of predictive models to a certain target measure is a desirable property of MLC approaches. Methods based on \emph{gradient boosting}, which guide the construction of an ensemble of weak learners towards the minimization of a given loss function, appear to be appealing with regard to this requirement. In fact, several boosting-based approaches for MLC have been proposed in the literature. Though many of these methods are restricted to the use of label-wise decomposable loss functions (e.g.,\ \cite{zhang2020}, \cite{si2017} or \cite{johnson2005}), including methods that focus on ranking losses (e.g.,\ \cite{jung2017}, \cite{dembczynski2012b} or \cite{schapire2000}), gradient boosting has also been used to minimize non-decomposable losses (e.g.,\ \cite{rapp2020} or \cite{amit2007}).

To be able to take dependencies between labels into account, problem transformation methods, such as \emph{Label Powerset}~\cite{tsoumakas2010}, \emph{RAKEL}~\cite{tsoumakas2007} or \emph{(Probabilistic) Classifier Chains}~\cite{cheng2010,read2009}, transform the original learning task into several sub-problems that can be solved by the means of binary classification algorithms. Compared to binary relevance, where each label is considered in isolation, these approaches come with high computational demands. To compensate for this, methods like \emph{HOMER}~\cite{tsoumakas2008}, \emph{Compressed Sensing}~\cite{zhou2012}, \emph{Canonical Correlation Analysis}~\cite{sun2010}, \emph{Principal Label Space Transformation}~\cite{tai2012} or \emph{Label Embeddings}~\cite{kumar2019,huang2017,bhatia2015} aim to reduce the complexity of the label space to be dealt with by multi-label classifiers. Notwithstanding that such a reduction in complexity is indispensable in cases where thousands or even millions of labels must be handled, it often remains unclear what measure such methods aim to optimize. In this work, we explicitly focus on the minimization of non-decomposable loss functions in cases where the original problem is tractable. We therefore aim at real-world problems with up to a few hundred labels, where such metrics, especially the Subset 0/1 loss, are considered as important quality measures.

As our contribution, we propose a novel method to be integrated into the gradient boosting framework. Based on the derivatives that guide the optimization process, it maps the labels to a predefined number of bins. If the loss function is non-decomposable, this reduction in dimensionality limits the computational efforts needed to evaluate potential weak learners. Unlike the reduction methods mentioned above, our approach dynamically adjusts to different regions in input space for which a learner may predict. Due to the exploitation of the derivatives, the impact of the approximation is kept at a minimum. We investigate the effects on training time and predictive performance using \emph{BOOMER}~\cite{rapp2020}, a boosting algorithm for learning multi-label rules. In general, the proposed method is not limited to rules and can easily be extended to gradient boosted decision trees.

\section{Preliminaries}
\label{sec:preliminaries}

In this section, we briefly recapitulate the multi-label classification setting and introduce the notation used in this work. We also discuss the methodology used to tackle multi-label problems by utilizing the gradient boosting framework.

\subsection{Multi-label Classification}
\label{sec:mlc}

We deal with multi-label classification as a supervised learning problem, where a model is fit to labeled training data $\dataset = \left \{ \left( \ex_{1}, \truelabelvect_{1} \right), \dots, \left( \ex_{\numex}, \truelabelvect_{\numex} \right) \right \} \subset \attrspace \times \labelspace$. Each example in such data set is characterized by a vector $\ex = \left( \attrval_{1}, \dots, \attrval_{\numattr} \right) \in \attrspace$ that assigns constant values to numerical or nominal attributes $\attr_{1}, \dots, \attr_{\numattr}$. In addition, an example may be associated with an arbitrary number of labels out of a predefined label set $\labelset = \left \{ \labl_{1}, \dots, \labl_{\numlabels} \right \}$. The information, whether individual labels are relevant ($1$) or irrelevant ($-1$) to a training example, is specified in the form of a binary label vector $\truelabelvect = \left( \truelabel_{1}, \dots, \truelabel_{\numlabels} \right) \in \labelspace$. The goal is to learn a model $\classifier: \attrspace \to \labelspace$ that maps any given example to a predicted label vector $\predvect = \left( \predlabel_{1}, \dots, \predlabel_{\numlabels} \right) \in \labelspace$. It should generalize beyond the given training examples such that it can be used to obtain predictions for unseen data.

Ideally, the training process can be tailored to a certain loss function such that the predictions minimize the expected risk with respect to that particular loss. In multi-label classification several meaningful loss functions with different characteristics exist. In the literature, one does usually distinguish between label-wise \emph{decomposable} loss functions, such as the Hamming loss (see, e.g., \cite{tsoumakas2010} for a definition of this particular loss function), and \emph{non-decomposable} losses. The latter are considered to be particularly difficult to minimize, as it is necessary to take interactions between the labels into account \cite{dembczynski2012a}. Among this kind of loss functions is the \emph{Subset 0/1 loss}, which we focus on in this work. Given true and predicted label vectors, it is defined as
\begin{equation}
\label{eq:subset_01_loss}
\loss_{\text{Subs.}} \left( \truelabelvect_{\iterex}, \predvect_{\iterex} \right) \coloneqq \llbracket \truelabelvect_{\iterex} \neq \predvect_{\iterex} \rrbracket,
\end{equation}
where $\llbracket x \rrbracket$ evaluates to $1$ or $0$, if the predicate $x$ is true or false, respectively. As a wrong prediction for a single label is penalized as much as predicting incorrectly for several labels, the minimization of the Subset 0/1 loss is very challenging. Due to its interesting properties and its prominent role in the literature, we consider it as an important representative of non-decomposable loss functions.

\subsection{Multivariate Gradient Boosting}
\label{sec:boosting}

We build on a recently proposed extension to the popular gradient boosting framework that enables to minimize decomposable, as well as non-decomposable, loss functions in a multi-label setting \cite{rapp2020}. Said approach aims at learning ensembles $\ensemble_{\numclassifiers} = \{ \classifier_{1}, \dots, \classifier_{\numclassifiers} \}$ that consist of several weak learners. In multi-label classification, each ensemble member can be considered as a predictive function that returns a vector of real-valued confidence scores
\begin{equation}
\label{eq:confvect}
\confvect_{\iterex}^{\iterclassifiers} = \classifier_{\iterclassifiers} \left( \ex_{\iterex} \right) = \left( \conf_{\iterex 1}^{\iterclassifiers}, \dots, \conf_{\iterex \numlabels}^{\iterclassifiers} \right) \in \mathbb{R}^{\numlabels}
\end{equation}
for any given example. Each confidence score $\conf_{\iterex \iterlabels}$ expresses a preference towards predicting the corresponding label $\labl_{\iterlabels}$ as relevant or irrelevant, depending on whether the score is positive or negative. To compute an ensemble's overall prediction, the vectors that are provided by its members are aggregated by calculating the element-wise sum
\begin{equation}
\label{eq:confvect_aggregated}
\confvect_{\iterex} = \ensemble_{\numclassifiers} \left( \ex_{\iterex} \right) = \confvect_{\iterex}^{1} + \dots + \confvect_{\iterex}^{\numclassifiers} \in \mathbb{R}^{\numlabels},
\end{equation}
which can be discretized in a second step to obtain a binary label vector.

An advantage of gradient boosting is the capability to tailor the training process to a certain (surrogate) loss function. Given a loss $\loss$, an ensemble should be trained such that the global objective
\begin{equation}
\label{eq:objective_global}
\objective \left( \ensemble_{\numclassifiers} \right) = \sum_{\iterex = 1}^{\numex} \loss \left( \truelabelvect_{\iterex}, \confvect_{\iterex} \right) + \sum_{\iterclassifiers = 1}^{\numclassifiers} \regterm \left( \classifier_{\iterclassifiers} \right)
\end{equation}
is minimized. The use of a suitable \emph{regularization term} $\regterm$ may help to avoid overfitting and to converge towards a global optimum, if the loss function is not convex.

Gradient boosting is based on constructing an ensemble of additive functions following an iterative procedure, where new ensemble members are added step by step. To direct the step-wise training process towards a model that optimizes the global objective in the limit, \eqref{eq:objective_global} is rewritten based on the derivatives of the loss function. Like many recent boosting-based approaches (e.g., \cite{chen2016}, \cite{ke2017} or \cite{zhang2020}), we rely on the second-order Taylor approximation. Given a loss function that is twice differentiable, this results in the stagewise objective function
\begin{equation}
\label{eq:objective_stagewise}
\approxobjective \left( \classifier_{\iterclassifiers} \right) = \sum_{\iterex = 1}^{\numex} \left( \gradientvect_{\iterex} \confvect_{\iterex}^{\iterclassifiers} + \frac{1}{2} \confvect_{\iterex}^{\iterclassifiers} \hessianmatr_{\iterex} \confvect_{\iterex}^{\iterclassifiers} \right) + \regterm \left( \classifier_{\iterclassifiers} \right),
\end{equation}
which should be minimized by the ensemble member that is added at the $\iterclassifiers$-th training iteration. The gradient vector $\gradientvect_{\iterex} = \left( \gradient_{\iterex \iterrows} \right)_{1 \leq \iterrows \leq \numlabels}$ consist of the first-order partial derivatives of $\loss$ with respect to predictions of the current model for an example $\ex_{\iterex}$ and labels $\labl_{1}, \dots, \labl_{\numlabels}$. Accordingly, the second-order partial derivatives form the Hessian matrix 
$\hessianmatr_{\iterex} = \left( \hessian_{\iterex \iterrows \itercols} \right)_{1 \leq \iterrows, \itercols \leq \numlabels}$. The individual gradients and Hessians are formally defined as
\begin{equation}
\gradient_{\iterrows}^{\iterex} = \frac{\partial \loss}{\partial \conf_{\iterex \iterrows}} \left( \truelabelvect_{\iterex},  \ensemble_{\iterclassifiers - 1} \left( \ex_{\iterex} \right) \right) \quad \text{and} \quad \hessian_{\iterrows \itercols}^{\iterex} = \frac{\partial \loss}{\partial \conf_{\iterex \iterrows} \partial \conf_{\iterex \itercols}} \left( \truelabelvect_{\iterex},  \ensemble_{\iterclassifiers - 1} \left( \ex_{\iterex} \right) \right).
\end{equation}

The confidence scores that are predicted by an ensemble member $\classifier_{\iterclassifiers}$ for individual labels must be chosen such that the stagewise objective in \eqref{eq:objective_stagewise} is minimized. To derive a formula for calculating the predicted scores, the partial derivative of \eqref{eq:objective_stagewise} with respect to the prediction for individual labels must be equated to zero. In case of a decomposable loss function, this results in a closed form solution that enables to compute the prediction for each label independently. In the general case, i.e., when the loss function is non-decomposable and the prediction should not be restricted to a single label, one obtains a system of $\numlabels$ linear equations
\begin{equation}
\label{eq:linear_system}
\left( \hessianmatr + \regmatr \right) \confvect = -\gradientvect.
\end{equation}

Whereas the elements of the Hessian matrix $\hessianmatr$ and the gradient vector $\gradientvect$ can be considered as coefficients and ordinates, the vector $\confvect$ consists of the unknowns to be determined. The matrix $\regmatr$ is used to take the regularization into account. In this work, we use the $L_2$ regularization term \begin{equation}
\label{eq:regterm}
\regterm_{\text{L2}} \left( \rul_{\iterclassifiers} \right) = \frac{1}{2} \regweight \left \lVert \confvect^{\iterclassifiers} \right \rVert_{2}^{2} \, ,
\end{equation}
where $\left \lVert \vec{x} \right \rVert_{2}$ is the Euclidean norm and $\regweight \geq 0$ controls the weight of the regularization. In this particular case, the regularization matrix $\regmatr = \diag \left( \regweight \right)$ is a diagonal matrix with the value $\regweight$ on the diagonal.

As the target function to be minimized, we use the logistic loss function
\begin{equation}
\label{eq:log_loss_example_wise}
\loss_{\text{ex.w.-log}} \left( \truelabelvect_{\iterex}, \confvect_{\iterex} \right) \coloneqq \log \left( 1 + \sum_{\iterlabels = 1}^{\numlabels} \exp \left( -\truelabel_{\iterex \iterlabels} \conf_{\iterex \iterlabels} \right) \right),
\end{equation}
which has previously been used with the BOOMER algorithm as a surrogate for the Subset 0/1 loss and was originally proposed by Amit~et~al.~\cite{amit2007}.

\subsection{Ensembles of Multi-label Rules}
\label{sec:rules}

We rely on multi-label rules as the individual building blocks of ensembles that are trained according to the methodology in \sect~\ref{sec:boosting}. In accordance with \eqref{eq:confvect}, each rule can be considered as a function
\begin{equation}
\rul \left( \ex \right) = \body \left( \ex \right) \head
\end{equation}
that predicts a vector of confidence scores for a given example.

The \emph{body} of a rule $\body: \attrspace \to \left \{ 0, 1 \right \}$ is a conjunction of one or several conditions that compare a given example's value for a particular attribute $\attr_{\iterattr}$ to a constant using a relational operator like $\leq$ and $>$, if the attribute is numerical, or $=$ and $\neq$, if it is nominal. If an example satisfies all conditions in the body, i.e., if $\body \left( \ex \right) = 1$, it is \emph{covered} by the respective rule. In such case, the scores that are contained in the rule's \emph{head} $\head \in \mathbb{R}^{\numlabels}$ are returned. It assigns a positive or negative confidence score to each label, depending on whether the respective label is expected to be mostly relevant or irrelevant to the examples that belong to the region of the input space $\attrspace$ that is covered by the rule. If an example is not covered, i.e., if $\body \left( \ex \right) = 0$, a null vector is returned. In such case, the rule does not have any effect on the overall prediction, as can be seen in \eqref{eq:confvect_aggregated}.

If individual elements in a rule's head are set to zero, the rule does not provide a prediction for the corresponding labels. The experimental results reported by Rapp~et~al.~\cite{rapp2020} suggest that \emph{single-label rules}, which only provide a non-zero prediction for a single label, tend to work well for minimizing decomposable loss functions. Compared to \emph{multi-label rules}, which jointly predict for several labels, the induction of such rules is computationally less demanding, as a closed form solution for determining the predicted scores exists. However, the ability of multi-label rules to express local correlations between several labels, which hold for the examples they cover, has been shown to be crucial when it comes to non-decomposable losses.

To construct a rule that minimizes \eqref{eq:objective_stagewise}, we conduct a top-down greedy search, as it is commonly used in inductive rule learning (see, e.g., \cite{fuernkranz2012} for an overview on the topic). Initially, the search starts with an empty body that does not contain any conditions and therefore is satisfied by all examples. By adding new conditions to the body, the rule is successively specialized, resulting in less examples being covered in the process. For each candidate rule that results from adding a condition, the confidence scores to be predicted for the covered examples are calculated by solving \eqref{eq:linear_system}. By substituting the calculated scores into \eqref{eq:objective_stagewise}, an estimate of the rule's quality is obtained. Among all possible refinements, the one that results in the greatest improvement in terms of quality is chosen. The search stops as soon as the rule cannot be improved by adding a condition.

Rules are closely related to the more commonly used decision trees, as each tree can be viewed as a set of non-overlapping rules. At each training iteration, a rule-based boosting algorithm focuses on a single region of the input space for which the model can be improved the most. In contrast, gradient boosted decision trees do always provide predictions for the entire input space. Due to their conceptual similarities, the ideas presented in this paper are not exclusive to rules, but can also be applied to decision trees.

\section{Gradient-based Label Binning}
\label{sec:binning}

In this section, we present \emph{Gradient-based Label Binning} (GBLB), a novel method that aims at reducing the computational costs of the multivariate boosting algorithm discussed in \sect~\ref{sec:boosting}. Although the method can be used with any loss function, it is intended for use cases where a non-decomposable loss should be minimized. This is, because it explicitly addresses the computational bottleneck of such training procedure --- the need to solve the linear system in \eqref{eq:linear_system} --- which reduces to an operation with linear complexity in the decomposable case.

\subsection{Complexity Analysis}
\label{sec:complexity}

The objective function in \eqref{eq:objective_stagewise}, each training iteration aims to minimize, depends on gradient vectors and Hessian matrices that correspond to individual training examples. Given $\numlabels$ labels, the former consist of $\numlabels$ elements, whereas the latter are symmetric matrices with $\numlabels \left( \numlabels + 1 \right) \mathbin{/} 2$ non-zero elements, one for each label, as well as for each pair of labels. The induction of a new rule, using a search algorithm as described in \sect~\ref{sec:rules}, requires to sum up the gradient vectors and Hessian matrices of the covered examples to form the linear system in \eqref{eq:linear_system}. Instead of computing the sums for each candidate rule individually, the candidates are processed in a predetermined order, such that each one covers one or several additional examples compared to its predecessor (see, e.g., \cite{mehta1996} for an early description of this idea). As a result, an update of the sums with complexity $\bigo \left( \numlabels^2 \right)$ must be performed for each example and attribute that is considered for making up new candidates.

\begin{algorithm}[t]
\SetKwInOut{Input}{input}\SetKwInOut{Output}{output}
  \Input{Gradient vector $\gradientvect$, Hessian matrix $\hessianmatr$, $L_2$ regularization weight $\regweight$}
  \Output{Predictions $\predvect$, quality score $s$, mapping $\vec{m}$ (if GBLB is used)}
  \vspace{4pt}
  \def\columnseprule{0.5pt}
  \begin{multicols}{2}
  Regularization matrix $\regmatr = \diag \left( \regweight \right)$ \\
  \nonl ~\\
  $\predvect = \textsc{dsysv} \left( -\gradientvect, \hessianmatr + \regmatr \right)$ \comment{cf.~\eqref{eq:linear_system}} \\ \label{ln:dsysv}
  $s = \textsc{ddot} \left( \predvect, \gradientvect \right) + \left( 0.5 \cdot \textsc{ddot} \left( \predvect, \textsc{dspmv} \left( \predvect, \hessianmatr \right) \right) \right)$ \comment{cf.~\eqref{eq:objective_stagewise}} \\ \label{ln:dspmv}
  \Return $\predvect, s$ \\
  \nonl Mapping $\vec{m} = \textsc{map\_to\_bins} \left( \gradientvect, \hessianmatr, \regweight \right)$ \\
  \nonl $\gradientvect, \hessianmatr, \regmatr = \textsc{aggregate} \left( \vec{m}, \gradientvect, \hessianmatr, \regweight \right)$ \\
  \nonl \centering$\vert$ \\
  \nonl \textit{same as left} \\
  \nonl $\vert$ \\
  \nonl \raggedright\Return $\predvect, s, \vec{m}$ \\
  \end{multicols}
  \vspace{1.2pt}
  \caption{Candidate evaluation without (left) / with GBLB (right)}
  \label{alg:evaluate_rule}
\end{algorithm}

\alg~\ref{alg:evaluate_rule} shows the steps that are necessary to compute the confidence scores to be predicted by an individual candidate rule, as well as a score that assesses its quality, if the loss function is non-decomposable. The modifications that are necessary to implement GBLB are shown to the right of the original lines of code. Originally, the given gradient vector and Hessian matrix, which result from summation over the covered examples, are used as a basis to solve the linear system in \eqref{eq:linear_system} using the Lapack routine \textsc{dsysv} (cf.~\alg~\ref{alg:evaluate_rule}, \algln~\ref{ln:dsysv}). The computation of a corresponding quality score by substituting the calculated scores into \eqref{eq:objective_stagewise}, involves invocations of the Blas operations \textsc{ddot} for vector-vector multiplication, as well as \textsc{dspmv} for vector-matrix multiplication (cf.~\alg~\ref{alg:evaluate_rule}, \algln~\ref{ln:dspmv}). Whereas the operation \textsc{ddot} comes with linear costs, the \textsc{dspmv} and \textsc{dsysv} routines have quadratic and cubic complexity, i.e., $\bigo \left( \numlabels^2 \right)$ and $\bigo \left( \numlabels^3 \right)$, respectively\footnote{Information on the complexity of the Blas and Lapack routines used in this work can be found at \url{http://www.netlib.org/lapack/lawnspdf/lawn41.pdf}.}. As \alg~\ref{alg:evaluate_rule} must be executed for each candidate rule, it is the computationally most expensive operation that takes part in a multivariate boosting algorithm aimed at the minimization of a non-decomposable loss function.

GBLB addresses the computational complexity of \alg~\ref{alg:evaluate_rule} by mapping the available labels to a predefined number of bins $\numbins$ and aggregating the elements of the gradient vector and Hessian matrix accordingly. If $\numbins \ll \numlabels$, this significantly reduces their dimensionality and hence limits the costs of the Blas and Lapack routines. As a result, given that the overhead introduced by the mapping and aggregation functions is small, we expect an overall reduction in training time.

In this work, we do not address the computational costs of summing up the gradients that correspond to individual examples. However, the proposed method has been designed such that it can be combined with methods that are dedicated to this aspect. Albeit restricting themselves to decomposable losses, Si~et~al.~\cite{si2017} have proposed a promising method that ensures that many gradients evaluate to zero. This approach, which was partly adopted by Zhang~and~Jung~\cite{zhang2020}, restricts the labels that must be considered to those with non-zero gradients. However, to maintain sparsity among the gradients, strict requirements must be fulfilled by the loss function. Among many others, the logistic loss function in \eqref{eq:log_loss_example_wise} does not meet these requirements. The approach that is investigated in this work does not impose any restrictions on the loss function.

\subsection{Mapping Labels to Bins}
\label{sec:mapping}

GBLB evolves around the idea of assigning the available labels $\labl_{1}, \dots, \labl_{\numlabels}$ to a predefined number of bins $\bin_{1}, \dots, \bin_{\numbins}$ whenever a potential ensemble member is evaluated during training (cf. \textsc{map\_to\_bins} in \alg~\ref{alg:evaluate_rule}). To obtain the index of the bin, a particular label $\lambda_{\iterlabels}$ should be assigned to, we use a mapping function $\mapfun: \mathbb{R} \rightarrow \mathbb{N}^{+}$ that depends on a given criterion $\crit_{\iterlabels} \in \mathbb{R}$. In this work, we use the criterion
\begin{equation}
\crit_{\iterlabels} = -\frac{\gradient_{\iterlabels}}{\hessian_{\iterlabels \iterlabels} + \regweight},
\end{equation}
which takes the gradient and Hessian for the respective label, as well as the $L_2$ regularization weight, into account. It corresponds to the optimal prediction when considering the label in isolation, i.e., when assuming that the predictions for other labels will be zero. As the criterion can be obtained for each label individually, the computational overhead is kept at a minimum.

Based on the assignments that are provided by a mapping function $\mapfun$, we denote the set of label indices that belong to the $\iterbins$-th bin as
\begin{equation}
\bin_{\iterbins} = \left \{ \iterlabels \in \left \{ 1, \dots, \numlabels \right \}\ \rvert\ \mapfun \left( \crit_{\iterlabels} \right) = \iterbins \right \}.
\end{equation}

Labels should be assigned to the same bin if the corresponding confidence scores, which will be presumably be predicted by an ensemble member, are close to each other. If the optimal scores to be predicted for certain labels are very different in absolute size or even differ in their sign, the respective labels should be mapped to different bins. Based on this premise, we limit the number of distinct scores, an ensemble member may predict, by enforcing the restriction
\begin{equation}
\label{eq:binning_equality}
\conf_{\iterrows} = \conf_{\itercols}, \forall \iterrows, \itercols \in \bin_{\iterbins}.
\end{equation}
It requires that a single score is predicted for all labels that have been assigned to the same bin. Given that the mentioned prerequisites are met, we expect the difference between the scores that are predicted for a bin and those that are optimal with respect to its individual labels to be reasonably small.

\subsection{Equal-width Label Binning}
\label{sec:equal_width}

Principally, different approaches to implement the mapping function $\mapfun$ are conceivable. We use \emph{equal-width} binning, as this well-known method provides two advantages: First, unlike other methods, such as equal-frequency binning, it does not involve sorting and can therefore be applied in linear time. Second, the boundaries of the bins are chosen such that the absolute difference between the smallest and largest value in a bin, referred to as the \emph{width}, is the same for all bins. As argued in \sect~\ref{sec:mapping}, this is a desirable property in our particular use case. Furthermore, we want to prevent labels, for which the predicted score should be negative, from being assigned to the same bin as labels, for which the prediction should be positive. Otherwise, the predictions would be suboptimal for some of these labels. We therefore strictly separate between \emph{negative} and \emph{positive bins}. Given $\numbinsneg$ negative and $\numbinspos$ positive bins, the width calculates as
\begin{equation}
\widthneg = \frac{\maxneg - \minneg}{\numbinsneg} \quad \text{and} \quad \widthpos = \frac{\maxpos - \minpos}{\numbinspos},
\end{equation}
for the positive and negative bins, respectively. By $\maxneg$ and $\maxpos$ we denote the largest value in $\left \{ \crit_{1}, \dots, \crit_{\numlabels} \right \}$ with negative and positive sign, respectively. Accordingly, $\minneg$ and $\minpos$ correspond to the smallest value with the respective sign. Labels for which $\crit_{\iterlabels} = 0$, i.e., labels with zero gradients, can be ignored. As no improvement in terms of the loss function can be expected, we explicitly set the prediction to zero in such case.

Once the width of the negative and positive bins has been determined, the mapping from individual labels to one of the $\numbins = \numbinsneg + \numbinspos$ bins can be obtained via the function
\begin{equation}
\label{eq:binfun}
\mapfun_{\text{eq.-width}} \left( \crit_{\iterlabels} \right) = 
\begin{cases}
    \min \left( \lfloor \frac{\crit_{\iterlabels} - \minneg}{\widthneg} \rfloor + 1, \numbinsneg \right),   & \text{if}\ \crit_{\iterlabels} < 0 \\
    \min \left( \lfloor \frac{\crit_{\iterlabels} - \minpos}{\widthpos} \rfloor + 1, \numbinspos \right) + \numbinsneg, & \text{if}\ \crit_{\iterlabels} > 0.
\end{cases}
\end{equation}

\subsection{Aggregation of Gradients and Hessians}
\label{sec:aggregation}

By exploiting the restriction introduced in \eqref{eq:binning_equality}, the gradients and Hessians that correspond to labels in the same bin can be aggregated to obtain a gradient vector and a Hessian matrix with reduced dimensions (cf. \textsc{aggregate} in \alg~\ref{alg:evaluate_rule}). To derive a formal description of this aggregation, we first rewrite the objective function \eqref{eq:objective_stagewise} in terms of sums instead of using vector and matrix multiplications. This results in the formula
\begin{equation}
\label{eq:objective_sums}
\approxobjective \left( \classifier_{\iterclassifiers} \right) = \sum_{\iterex = 1}^{\numex} \sum_{\iterrows = 1}^{\numlabels} \left( \gradient_{\iterrows}^{\iterex} \conf_{\iterrows} + \frac{1}{2} \conf_{\iterrows} \left( \hessian_{\iterrows \iterrows}^{\iterex} \conf_{\iterrows} + \sum_{\substack{\itercols = 1,\\ \itercols \neq \iterrows}}^{\numlabels} \hessian_{\iterrows \itercols}^{\iterex} \conf_{\itercols} \right) \right) + \regterm \left( \classifier_{\iterclassifiers} \right).
\end{equation}

\begin{figure}[t]
\centering
\includegraphics[width=0.95\textwidth]{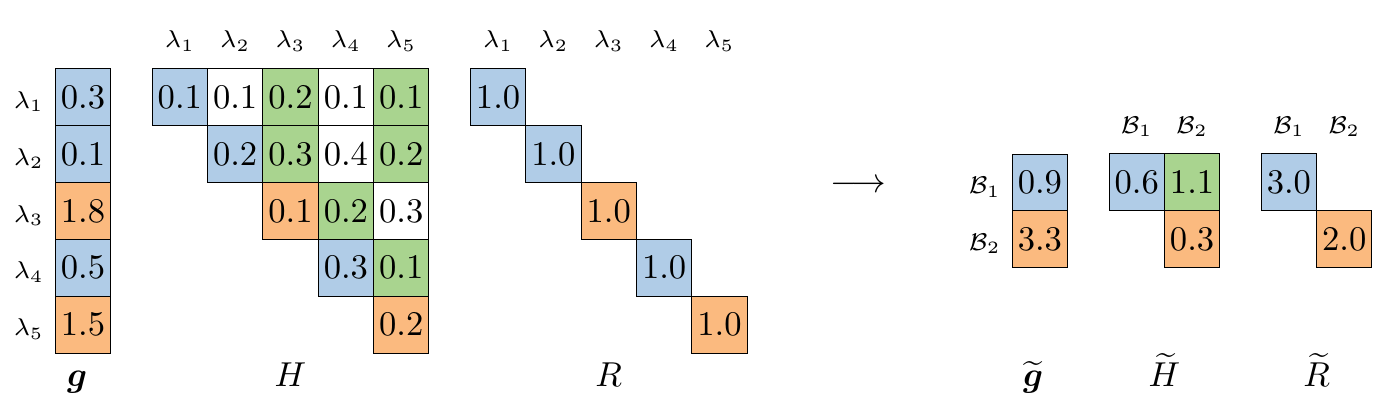}
\caption{Illustration of how a gradient vector, a Hessian matrix and a regularization matrix for five labels $\labl_{1}, \dots, \labl_{5}$ are aggregated with respect to two bins $\bin_{1} = \left \{ 1, 2, 4 \right \}$ and $\bin_{2} = \left \{ 3, 5 \right \}$ when using $L_2$ regularization with $\regweight = 1$. Elements with the same color are added up for aggregation.}
\label{fig:example}
\end{figure}

Based on the constraint given in \eqref{eq:binning_equality} and due to the distribution property of the multiplication, the equality
\begin{equation}
\label{eq:lemma}
\sum_{\iterrows = 1}^{\numlabels} x_{\iterrows} \conf_{\iterrows} = \sum_{\itercols = 1}^{\numbins} \left( \conf_{\itercols} \sum_{\iterrows \in \bin_{\itercols}} x_{\iterrows} \right),
\end{equation}
where $x_{\iterrows}$ is any term dependent on $\iterrows$, holds. It can be used to rewrite \eqref{eq:objective_sums} in terms of sums over the bins, instead of sums over the individual labels. For brevity, we denote the sum of the gradients, as well as the sum of the elements on the diagonal of the Hessian matrix, that correspond to the labels in bin $\bin_{\iterbins}$ as
\begin{equation}
\label{eq:aggregation1}
\bingradients_{\iterbins} = \sum_{\iterrows \in \bin_{\iterbins}} \gradient_{\iterrows} \quad \text{and} \quad \binhessians_{\iterbins \iterbins} = \sum_{\iterrows \in \bin_{\iterbins}} \hessian_{\iterrows \iterrows}.
\end{equation}

To abbreviate the sum of Hessians that correspond to a pair of labels that have been assigned to different bins $\bin_{\iterbins}$ and $\bin_{q}$, we use the short-hand notation
\begin{equation}
\label{eq:aggregation2}
\binhessians_{\iterbins q} = \sum_{\iterrows \in \bin_{\iterbins}} \sum_{\itercols \in \bin_{q}} \hessian_{\iterrows \itercols}.
\end{equation}

By exploiting \eqref{eq:lemma} and using the abbreviations introduced above, the objective function in \eqref{eq:objective_sums} can be rewritten as
\begin{equation}
\label{eq:objective_binning}
\approxobjective \left( \classifier_{\iterclassifiers} \right) = \sum_{\iterex = 1}^{\numex}  \sum_{\iterbins = 1}^{\numbins} \left( \conf_{\iterbins} \bingradients_{\iterbins}^{\iterex} + \frac{1}{2} \conf_{\iterbins} \left( \conf_{\iterbins} \binhessians_{\iterbins}^{\iterex} + \sum_{\substack{q = 1,\\ q \neq \iterbins}}^{\numbins} \conf_{q} \binhessians_{\iterbins q}^{\iterex} \right) \right) + \regterm \left( \classifier_{\iterclassifiers} \right),
\end{equation}
which can afterwards be turned into the original notation based on vector and matrix multiplications. The resulting formula
\begin{equation}
\approxobjective \left( \classifier_{\iterclassifiers} \right) = \sum_{\iterex = 1}^{\numex} \left( \bingradientvect_{\iterex} \confvect_{\iterex}^{\iterclassifiers} + \frac{1}{2} \confvect_{\iterex}^{\iterclassifiers} \binhessianmatr_{\iterex} \confvect_{\iterex}^{\iterclassifiers} \right) + \regterm \left( \classifier_{\iterclassifiers} \right)
\end{equation}
has the same structure as originally shown in \eqref{eq:objective_stagewise}. However, the gradient vector $\gradientvect$ and the Hessian matrix $\hessianmatr$ have been replaced by $\bingradientvect$ and $\binhessianmatr$, respectively. Consequently, when calculating the scores to be predicted by an ensemble member by solving \eqref{eq:linear_system}, the coefficients and ordinates that take part in the linear system do not correspond to individual labels, but result from the sums in \eqref{eq:aggregation1} and \eqref{eq:aggregation2}. As a result, number of linear equations has been reduced from the number of labels $\numlabels$ to the number of non-empty bins, which is at most $\numbins$.

An example that illustrates the aggregation of a gradient vector and a Hessian matrix is given in \fig~\ref{fig:example}. It also takes into account how the regularization matrix $\regmatr$ is affected. When dealing with bins instead of individual labels, the $L_2$ regularization term in \eqref{eq:regterm} becomes
\begin{equation}
\regterm_{\text{L2}} \left( \rul_{\iterclassifiers} \right) = \frac{1}{2} \regweight \sum_{\iterbins = 1}^{\numbins} \left( \left| \bin_{\iterbins} \right| \conf_{\iterbins}^2 \right),
\end{equation}
where $\left| \bin_{\iterbins} \right|$ denotes the number of labels that belong to a particular bin. As a consequence, the regularization matrix becomes $\binregmatr = \diag \left(\regweight \left| \bin_{1} \right|, \dots, \regweight \left| \bin_{\numbins} \right| \right)$.

\begin{table}[t]
\caption{Average training times (in seconds) per cross validation fold on different data sets (the number of labels is given in parentheses). The small numbers specify the speedup that results from using GBLB with the number of bins set to 32, 16, 8 and 4\% of the labels, or using two bins. Variants that are equivalent to two bins are omitted.}
\label{tab:evaluation_time}
\centering
\begin{tabular}{|l r|r|r r r r r r r r r r|}
\cline{3-13}
\multicolumn{2}{l|}{} & \multicolumn{1}{c|}{No} & \multicolumn{10}{c|}{GBLB} \\
\multicolumn{2}{l|}{} & \multicolumn{1}{c|}{GBLB} & \multicolumn{2}{c}{32\%}
                      & \multicolumn{2}{c}{16\%}
                      & \multicolumn{2}{c}{8\%}
                      & \multicolumn{2}{c}{4\%}
                      & \multicolumn{2}{c|}{2 bins} \\
\hline
Eurlex-sm & $\scriptstyle (201)$ & $46947$
                               & $54985$ & $\scriptstyle 0.85$
                               & $44872$ & $\scriptstyle 1.05$
                               & $38222$ & $\scriptstyle 1.23$
                               & $33658$ & $\scriptstyle 1.39$
                               & $\mathbf{21703}$ & $\scriptstyle 2.16$ \\
EukaryotePseAAC & $\scriptstyle (22)$ & $16033$
                               & $3593$ & $\scriptstyle 4.46$
                               & $2492$ & $\scriptstyle 6.43$
                               & $2195$ & $\scriptstyle 7.30$
                               & \multicolumn{2}{c}{---}
                               & $\mathbf{1534}$ & $\scriptstyle 10.45$ \\
Reuters-K500 & $\scriptstyle (103)$ & $12093$
                               & $6930$ & $\scriptstyle 1.75$
                               & $4197$ & $\scriptstyle 2.88$
                               & $3353$ & $\scriptstyle 3.61$
                               & $2803$ & $\scriptstyle 4.31$
                               & $\mathbf{2743}$ & $\scriptstyle 4.41$ \\
Bibtex & $\scriptstyle (159)$ & $2507$
                               & $2599$ & $\scriptstyle 0.96$
                               & $2765$ & $\scriptstyle 0.91$
                               & $2649$ & $\scriptstyle 0.95$
                               & $2456$ & $\scriptstyle 1.02$
                               & $\mathbf{2125}$ & $\scriptstyle 1.18$ \\
Yeast & $\scriptstyle (14)$ & $2338$
                               & $998$ & $\scriptstyle 2.34$
                               & $761$ & $\scriptstyle 3.07$
                               & $525$ & $\scriptstyle 4.45$
                               & \multicolumn{2}{c}{---}
                               & $\mathbf{521}$ & $\scriptstyle 4.49$ \\
Birds & $\scriptstyle (19)$ & $2027$
                               & $701$ & $\scriptstyle 2.89$
                               & $505$ & $\scriptstyle 4.01$
                               & $337$ & $\scriptstyle 6.01$
                               & \multicolumn{2}{c}{---}
                               & $\mathbf{336}$ & $\scriptstyle 6.03$ \\
Yahoo-Social & $\scriptstyle (39)$ & $1193$
                               & $261$ & $\scriptstyle 4.57$
                               & $217$ & $\scriptstyle 5.50$
                               & $192$ & $\scriptstyle 6.21$
                               & $\mathbf{139}$ & $\scriptstyle 8.58$
                               & $175$ & $\scriptstyle 6.82$ \\
Yahoo-Computers & $\scriptstyle (33)$ & $874$
                               & $172$ & $\scriptstyle 5.08$
                               & $134$ & $\scriptstyle 6.52$
                               & $126$ & $\scriptstyle 6.94$
                               & $\mathbf{101}$ & $\scriptstyle 8.65$
                               & $123$ & $\scriptstyle 7.11$ \\
Yahoo-Science & $\scriptstyle (40)$ & $735$
                               & $200$ & $\scriptstyle 3.67$
                               & $160$ & $\scriptstyle 4.59$
                               & $135$ & $\scriptstyle 5.44$
                               & $\mathbf{106}$ & $\scriptstyle 6.93$
                               & $136$ & $\scriptstyle 5.40$ \\
Yahoo-Reference & $\scriptstyle (33)$ & $571$
                               & $174$ & $\scriptstyle 3.28$
                               & $141$ & $\scriptstyle 4.05$
                               & $129$ & $\scriptstyle 4.43$
                               & $\mathbf{110}$ & $\scriptstyle 5.19$
                               & $137$ & $\scriptstyle 4.17$ \\
Slashdot & $\scriptstyle (20)$ & $518$
                               & $154$ & $\scriptstyle 3.36$
                               & $117$ & $\scriptstyle 4.43$
                               & $\mathbf{86}$ & $\scriptstyle 6.02$
                               & \multicolumn{2}{c}{---}
                               & $119$ & $\scriptstyle 4.35$ \\
EukaryoteGO & $\scriptstyle (22)$ & $191$
                               & $79$ & $\scriptstyle 2.42$
                               & $74$ & $\scriptstyle 2.58$
                               & $\mathbf{60}$ & $\scriptstyle 3.18$
                               & \multicolumn{2}{c}{---}
                               & $64$ & $\scriptstyle 2.98$ \\
Enron & $\scriptstyle (53)$ & $181$
                               & $69$ & $\scriptstyle 2.62$
                               & $52$ & $\scriptstyle 3.48$
                               & $48$ & $\scriptstyle 3.77$
                               & $47$ & $\scriptstyle 3.85$
                               & $\mathbf{44}$ & $\scriptstyle 4.11$ \\
Medical & $\scriptstyle (45)$ & $170$
                               & $60$ & $\scriptstyle 2.83$
                               & $57$ & $\scriptstyle 2.98$
                               & $55$ & $\scriptstyle 3.09$
                               & $\mathbf{50}$ & $\scriptstyle 3.40$
                               & $51$ & $\scriptstyle 3.33$ \\
Langlog & $\scriptstyle (75)$ & $132$
                               & $126$ & $\scriptstyle 1.05$
                               & $112$ & $\scriptstyle 1.18$
                               & $105$ & $\scriptstyle 1.26$
                               & $\mathbf{101}$ & $\scriptstyle 1.31$
                               & $102$ & $\scriptstyle 1.29$ \\
\hline
Avg. Speedup & & & \multicolumn{2}{r}{$\scriptstyle2.81$}& \multicolumn{2}{r}{$\scriptstyle3.58$}& \multicolumn{2}{r}{$\scriptstyle4.61$}& \multicolumn{2}{r}{$\scriptstyle\mathbf{4.86}$}& \multicolumn{2}{r|}{$\scriptstyle4.00$}\\
\hline
\end{tabular}
\end{table}

\section{Evaluation}
\label{sec:evaluation}

To investigate in isolation the effects GBLB has on predictive performance and training time, we chose a single configuration of the BOOMER algorithm as the basis for our experiments. We used 10-fold cross validation to train models that are aimed at the minimization of the Subset 0/1 loss on commonly used benchmark data sets\footnote{All data sets are available at \url{https://www.uco.es/kdis/mllresources}}. Each model consists of $5.000$ rules that have been learned on varying subsets of the training examples, drawn with replacement. The refinement of rules has been restricted to random subsets of the available attributes. As the learning rate and the $L_2$ regularization weight, we used the default values $0.3$ and $1.0$, respectively. Besides the original algorithm proposed in \cite{rapp2020}, we tested an implementation that makes use of GBLB\footnote{An implementation is available at \url{https://www.github.com/mrapp-ke/Boomer}}. For a broad analysis, we set the maximum number of bins to $32$, $16$, $8$, and $4\%$ of the available labels. In addition, we investigated an extreme setting with two bins, where all labels with positive and negative criteria are assigned to the same bin, respectively.

\begin{table}[t]
\caption{Predictive performance of different approaches in terms of the Subset 0/1 loss and the Hamming loss (smaller values are better).}
\label{tab:evaluation_performance}
\centering
\begin{tabular}{|l|l|r|r|r r r r r|}
\cline{3-9}
\multicolumn{2}{l|}{} & \multicolumn{1}{c|}{Label-} & \multicolumn{1}{c|}{No} & \multicolumn{5}{c|}{GBLB} \\
\multicolumn{2}{l|}{} & \multicolumn{1}{c|}{wise}   & \multicolumn{1}{c|}{GBLB} & 32\%
                                                    & 16\%
                                                    & 8\%
                                                    & 4\%
                                                    & 2 bins \\
\hline
\parbox[c]{3mm}{\multirow{15}{*}{\rotatebox[origin=c]{90}{Subset 0/1 loss}}}
 & Eurlex-sm & $61.63$ & $69.53$
                     & $45.07$
                     & $\mathbf{45.03}$
                     & $45.30$
                     & $45.08$
                     & $47.32$ \\
 & EukaryotePseAAC & $85.09$ & $65.43$
                     & $65.37$
                     & $\mathbf{65.28}$
                     & $65.68$
                     & \multicolumn{1}{c}{---}
                     & $65.52$  \\
 & Reuters-K500 & $71.37$ & $71.07$
                     & $53.70$
                     & $53.22$
                     & $53.07$
                     & $\mathbf{52.90}$
                     & $53.40$ \\
 & Bibtex & $85.99$ & $81.31$
                     & $\mathbf{77.28}$
                     & $77.44$
                     & $77.55$
                     & $77.32$
                     & $78.95$ \\
 & Yeast & $84.94$ & $76.54$
                     & $76.91$
                     & $76.42$
                     & $76.87$
                     & \multicolumn{1}{c}{---}
                     & $\mathbf{76.21}$  \\
 & Birds & $45.29$ & $45.30$
                     & $\mathbf{45.14}$
                     & $45.45$
                     & $45.60$
                     & \multicolumn{1}{c}{---}
                     & $46.53$  \\
 & Yahoo-Social & $50.65$ & $64.30$
                     & $34.49$
                     & $\mathbf{34.40}$
                     & $34.84$
                     & $35.47$
                     & $35.37$ \\
 & Yahoo-Computers & $58.04$ & $\mathbf{46.26}$
                     & $46.65$
                     & $47.09$
                     & $46.94$
                     & $47.70$
                     & $47.42$ \\
 & Yahoo-Science & $74.00$ & $85.80$
                     & $\mathbf{50.89}$
                     & $51.20$
                     & $52.07$
                     & $52.79$
                     & $52.04$ \\
 & Yahoo-Reference & $58.19$ & $74.14$
                     & $\mathbf{39.82}$
                     & $40.16$
                     & $40.73$
                     & $40.51$
                     & $40.48$ \\
 & Slashdot & $63.88$ & $\mathbf{46.62}$
                     & $46.64$
                     & $46.64$
                     & $47.73$
                     & \multicolumn{1}{c}{---}
                     & $47.22$  \\
 & EukaryoteGO & $30.63$ & $28.35$
                     & $28.39$
                     & $28.24$
                     & $\mathbf{28.10}$
                     & \multicolumn{1}{c}{---}
                     & $28.55$  \\
 & Enron & $88.19$ & $83.14$
                     & $83.32$
                     & $83.32$
                     & $83.38$
                     & $\mathbf{82.91}$
                     & $82.97$ \\
 & Medical & $28.25$ & $28.82$
                     & $23.13$
                     & $\mathbf{22.62}$
                     & $23.08$
                     & $23.23$
                     & $22.77$ \\
 & Langlog & $79.59$ & $78.84$
                     & $79.11$
                     & $79.25$
                     & $\mathbf{78.63}$
                     & $79.45$
                     & $79.45$ \\
\hline
\parbox[c]{3mm}{\multirow{15}{*}{\rotatebox[origin=c]{90}{Hamming loss}}}
 & Eurlex-sm & $0.55$ & $0.91$
                     & $0.40$
                     & $\mathbf{0.39}$
                     & $0.40$
                     & $0.40$
                     & $0.42$ \\
 & EukaryotePseAAC & $\mathbf{5.02}$ & $5.65$
                     & $5.64$
                     & $5.63$
                     & $5.67$
                     & \multicolumn{1}{c}{---}
                     & $5.66$  \\
 & Reuters-K500 & $1.11$ & $1.71$
                     & $1.11$
                     & $\mathbf{1.09}$
                     & $\mathbf{1.09}$
                     & $\mathbf{1.09}$
                     & $1.10$ \\
 & Bibtex & $\mathbf{1.25}$ & $1.45$
                     & $1.27$
                     & $1.27$
                     & $1.27$
                     & $1.28$
                     & $1.31$ \\
 & Yeast & $19.75$ & $19.01$
                     & $18.87$
                     & $19.08$
                     & $19.01$
                     & \multicolumn{1}{c}{---}
                     & $\mathbf{18.80}$  \\
 & Birds & $3.91$ & $3.79$
                     & $3.80$
                     & $3.79$
                     & $\mathbf{3.73}$
                     & \multicolumn{1}{c}{---}
                     & $3.87$  \\
 & Yahoo-Social & $1.90$ & $3.81$
                     & $\mathbf{1.79}$
                     & $1.80$
                     & $1.83$
                     & $1.87$
                     & $1.87$ \\
 & Yahoo-Computers & $3.10$ & $\mathbf{2.97}$
                     & $3.00$
                     & $3.02$
                     & $3.03$
                     & $3.08$
                     & $3.06$ \\
 & Yahoo-Science & $2.83$ & $5.85$
                     & $\mathbf{2.74}$
                     & $2.75$
                     & $2.81$
                     & $2.84$
                     & $2.79$ \\
 & Yahoo-Reference & $2.30$ & $4.95$
                     & $\mathbf{2.28}$
                     & $2.30$
                     & $2.34$
                     & $2.33$
                     & $2.32$ \\
 & Slashdot & $\mathbf{4.02}$ & $4.24$
                     & $4.24$
                     & $4.25$
                     & $4.37$
                     & \multicolumn{1}{c}{---}
                     & $4.30$  \\
 & EukaryoteGO & $\mathbf{1.89}$ & $1.95$
                     & $1.95$
                     & $1.94$
                     & $1.92$
                     & \multicolumn{1}{c}{---}
                     & $1.98$  \\
 & Enron & $\mathbf{4.53}$ & $4.72$
                     & $4.77$
                     & $4.77$
                     & $4.72$
                     & $4.72$
                     & $4.73$ \\
 & Medical & $0.84$ & $1.05$
                     & $0.80$
                     & $\mathbf{0.77}$
                     & $0.79$
                     & $0.81$
                     & $0.79$ \\
 & Langlog & $1.52$ & $1.52$
                     & $\mathbf{1.50}$
                     & $1.51$
                     & $\mathbf{1.50}$
                     & $1.52$
                     & $1.52$ \\
\hline
\end{tabular}
\end{table}

\tab~\ref{tab:evaluation_time} shows the average time per cross validation fold that is needed by the considered approaches for training. Compared to the baseline that does not use GBLB, the training time can always be reduced by utilizing GBLB with a suitable number of bins. Using fewer bins tends to speed up the training process, although approaches that use the fewest bins are not always the fastest ones. On average, limiting the number of bins to $4\%$ of the labels results in the greatest speedup (by factor 5). However, the possible speedup depends on the data set at hand. E.g., on the data set ``EukaryotePseAac'' the average training time is reduced by factor $10$, whereas no significant speedup is achieved for ``Bibtex''.

\begin{figure}[t]
\centering
\includegraphics[width=0.95\textwidth]{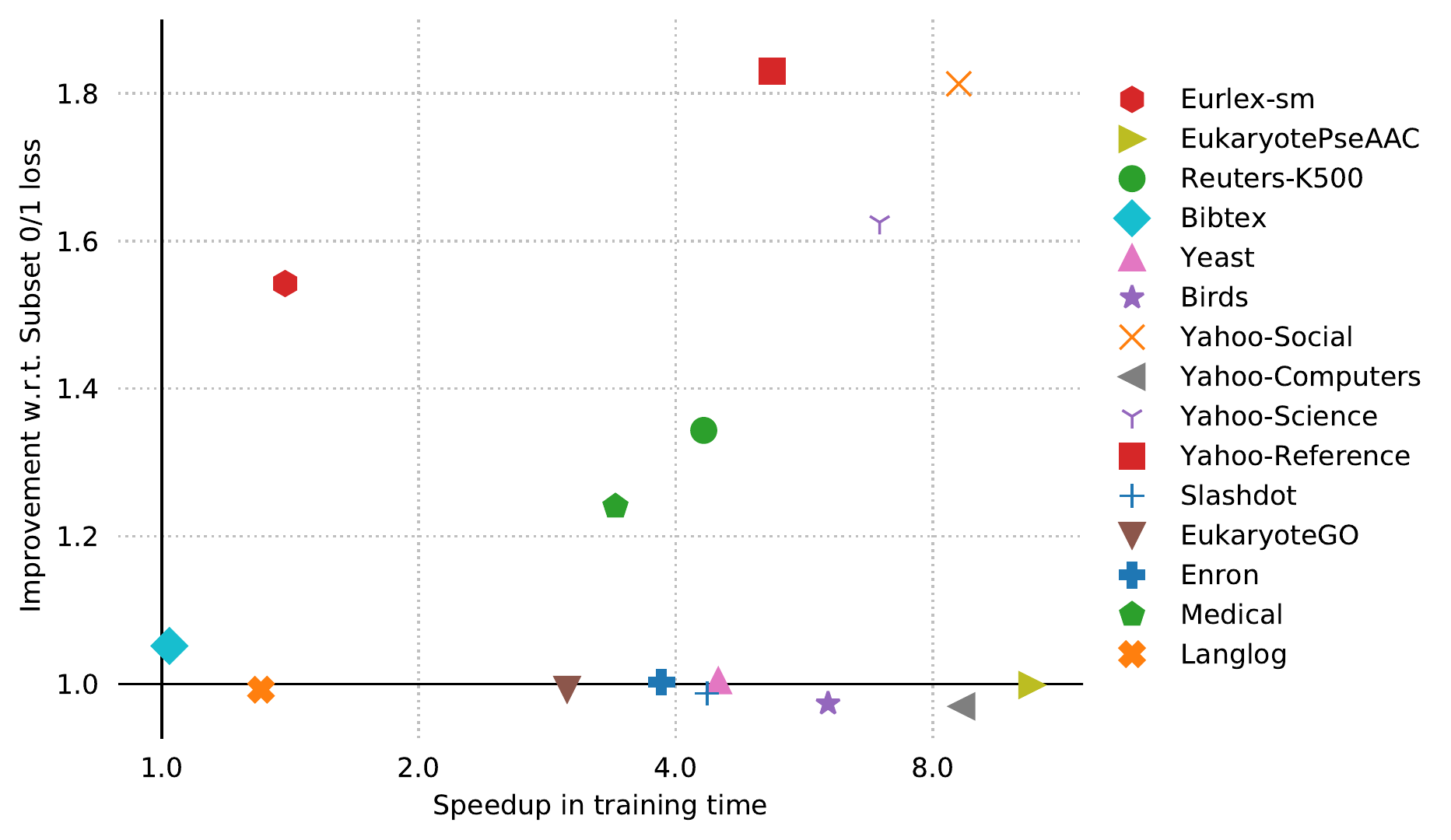}
\caption{Relative difference in training time and Subset 0/1 loss (both calculated as the baseline's value divided by the value of the respective approach) per cross validation fold that results from using GBLB with the number of bins set to 4\% of the labels.}
\label{fig:tradeoff}
\end{figure}

To be useful in practice, the speedup that results from GBLB should not come with a significant deterioration in terms of the target loss. We therefore report the predictive performance of the considered approaches in \tab~\ref{tab:evaluation_performance}. Besides the Subset 0/1 loss, which we aim to minimize in this work, we also include the Hamming loss as a commonly used representative of decomposable loss functions. When focusing on the Subset 0/1 loss, we observe that the baseline algorithm without GBLB exhibits subpar performance on some data sets, namely ``Eurlex-sm'', ``Reuters-K500'', ``Bibtex'', ``Yahoo-Social'', ``Yahoo-Science'', ``Yahoo-Reference'' and ``Medical''. This becomes especially evident when compared to an instantiation of the algorithm that targets the Hamming loss via minimization of a label-wise decomposable logistic loss function (cf.~\cite{rapp2020}, Eq.~6). In said cases, the latter approach performs better even though it is not tailored to the Subset 0/1 loss. Although the baseline performance could most probably be improved by tuning the regularization weight, we decided against parameter tuning, as it exposes an interesting property of GBLB. On the mentioned data sets, approaches that use GBLB appear to be less prone to converge towards local minima. Regardless of the number of bins, they clearly outperform the baseline. According to the Friedman test, these differences are significant with $\alpha = 0.01$. The Nemenyi post-hoc test yields critical distances for each of the GBLB-based approaches, when compared to the baseline. On the remaining data sets, where the baseline without GBLB already performs well, the use of GBLB produces competitive results. In these cases, the Friedman test confirms the null hypothesis with $\alpha = 0.1$. An overview of how the training time and the predictive performance in terms of the Subset 0/1 loss is affected, when restricting the number of bins to $4\%$ of the labels, is given in \fig~\ref{fig:tradeoff}.

\begin{figure}[t]
\centering
\includegraphics[width=1.0\textwidth]{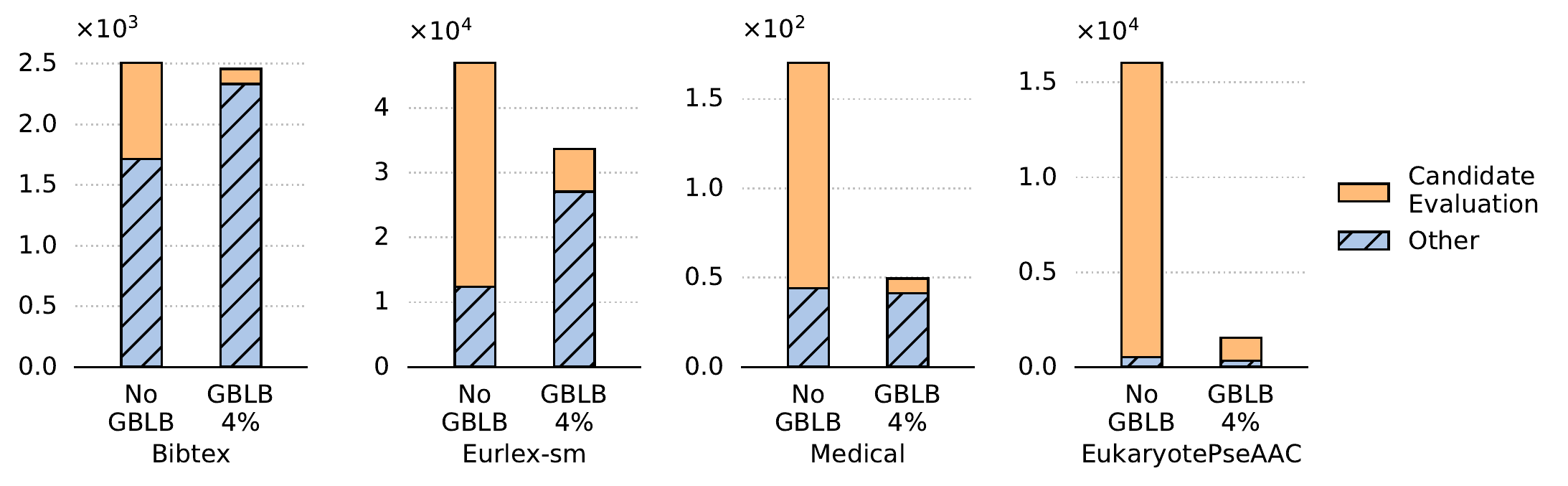}
\caption{Average proportion of training time per cross validation fold that is used for the evaluation of candidate rules without GBLB and when using GBLB with the number of bins set to 4\% of the labels.}
\label{fig:proportion}
\end{figure}

To better understand the differences in speedups that may be achieved by using GBLB, a detailed analysis is given in the following for four data sets with varying characteristics. In \fig~\ref{fig:proportion}, we depict the training time that is needed by the baseline approach, as well as by a GBLB-based approach with the number of bins set to $4\%$ of the labels. Besides the total training time, we also show the amount of time spent on the evaluation of candidate rules (cf.~\alg~\ref{alg:evaluate_rule}), which is the algorithmic aspect addressed by GBLB. For all given scenarios, it can be seen that the time needed for candidate evaluation could successfully be reduced. Nevertheless, the effects on the overall training time vastly differ. On the data sets ``Bibtex'' and `Eurlex-sm'', the time spent on parts of the algorithm other than the candidate evaluation increased when using GBLB, which is a result of more specific rules being learned. On the one hand, this required more candidates to be evaluated and therefore hindered the overall speedup. On the other hand, the resulting rules clearly outperformed the baseline according to \tab~\ref{tab:evaluation_performance}. On the data set ``Bibtex'', even without GBLB, the candidate evaluation was not the most expensive aspect of training. Due to its binary attributes, the number of potential candidates is small compared to the large number of examples. As a result, most of the computation time is spent on summing up the gradients and Hessians of individual examples (cf.~\sect~\ref{sec:complexity}). The impact of speeding up the candidate evaluation is therefore limited. On the data sets ``Medical'' and ``EukaryotePseAAC'', where the candidate evaluation was the most expensive aspect to begin with, a significant reduction of training time could be achieved by making that particular operation more efficient. The time spent on other parts of the algorithm remained mostly unaffected in these cases. As mentioned earlier, this includes the summation of gradients and Hessians, which becomes the most time consuming operation when using GBLB. Addressing this aspect holds the greatest potential for further performance improvements.

\section{Conclusion}
\label{sec:conclusion}

In this work we presented a novel approximation technique for use in multivariate boosting algorithms. Based on the derivatives that guide the training process, it dynamically assigns the available labels to a predefined number of bins. Our experiments, based on an existing rule learning algorithm, confirm that this reduction in dimensionality successfully reduces the training time that is needed for minimizing non-decomposable loss functions, such as the Subset 0/1 loss. According to our results, this speedup does not come with any significant loss in predictive performance. In several cases the proposed method even outperforms the baseline by a large extend due to its ability to overcome local minima without the necessity for extensive parameter tuning.

Despite our promising results, the use of non-decomposable loss functions in the boosting framework remains computationally challenging. Based on the analysis in this paper, we plan to extend our methodology with the ability to exploit sparsity in the label space. When combined with additional measures, the proposed method could become an integral part of more efficient algorithms that are capable of natively minimizing non-decomposable loss functions.

\subsubsection*{Acknowledgments}

This work was supported by the German Research Foundation (DFG) under grant number 400845550. Computations were conducted on the Lichtenberg high performance computer of the TU Darmstadt.

\bibliography{bibliography}
\bibliographystyle{splncs04}

\end{document}